\documentclass{article}
\usepackage[utf8]{inputenc}
\usepackage[round]{natbib}
\usepackage{gb4e}
\usepackage{paralist}
\usepackage{hyperref}
\usepackage{authblk}

\usepackage{linguex}
\usepackage{url}

\title{On the proper role of linguistically-oriented deep net analysis in linguistic theorizing}
\author[1,2]{Marco Baroni}
\affil[1]{Universitat Pompeu Fabra}
\affil[2]{Catalan Institution for Research and Advanced Studies (ICREA)}

\begin{document}

\maketitle

\begin{abstract}
A lively research field has recently emerged that uses experimental
methods to probe the linguistic behavior of modern deep
networks. While work in this tradition often reports intriguing
results about the grammatical skills of deep nets, it is not clear
what their implications for linguistic theorizing should be. As a
consequence, linguistically-oriented deep net analysis has had very
little impact on linguistics at large. In this chapter, I suggest that
deep networks should be treated as theories making explicit
predictions about the acceptability of linguistic utterances. I argue
that, if we overcome some obstacles standing in the way of
seriously pursuing this idea, we will gain a powerful new theoretical
tool, complementary to mainstream algebraic approaches.

\end{abstract}

\section{Introduction}\label{intro}

During the last decade, deep neural networks have come to dominate the
field of natural language processing (NLP)
\citep{Sutskever:etal:2014,Vaswani:etal:2017,Devlin:etal:2019}. While
earlier approaches to NLP relied on tools, such as part-of-speech
taggers and parsers, that extracted linguistic knowledge from explicit
manual annotation of text corpora \citep{Jurafsky:Martin:2008},
deep-learning-based methods typically adopt an ``end-to-end''
approach: A deep net is directly trained to associate some form of
natural linguistic input (e.g., text in a language) to a corresponding
linguistic output (e.g., the same text in a different language),
dispensing with the traditional pipeline of intermediate linguistic
modules and the related annotation of latent linguistic structure
(e.g., syntactic parses of source and target sentences)
\citep{Goldberg:2017,Lappin:2021}.

This paradigm shift has important implications for the relation
between theoretical and computational linguistics. The issue of which
linguistic formalisms might provide the best annotation schemes to
develop effective NLP tools is no longer relevant. Instead, the last
few years have seen the raise of a new field of investigation
consisting in the experimental analysis of the grammatical skills of
deep nets trained without the injection of any explicit linguistic
knowledge. For the remainder of the chapter, I will refer to this
research area as LODNA, for \emph{linguistically-oriented deep net
  analysis}. LODNA takes the perspective of a
psycholinguist \citep{Futrell:etal:2019}, or perhaps more accurately
that of an ethologist \citep{McCloskey:1991,Scholte:2018}, designing
sophisticated experiments to ``probe'' the knowledge implicit in a species' behavior.

LODNA is currently a very active research field, with many papers
focusing on whether neural networks have correctly induced specific
kinds of grammatical generalization
\citep[e.g.,][]{Linzen:etal:2016,Chowdhury:Zamparelli:2018,Futrell:etal:2019,Chaves:2020},
as well as benchmarks attempting to probe their linguistic competence
at multiple levels \citep{Conneau:etal:2018,Warstadt:etal:2019}. LODNA
papers account for a significant proportion of the work presented at
annual events such as the Society for Computation in Linguistics
conference and the BlackBox NLP workshop.

LODNA is well-motivated from a machine-learning perspective.
Understanding how a system behaves is a prerequisite to improving it,
and it is important in the perspective of AI safety and
explainability \citep{Belinkov:Glass:2019,Xie:etal:2020}. However,
there is no doubt that the grammatical performance of deep nets is
also extremely intriguing from a linguistic perspective, particularly
because the architectural primitives of these models (such as
distributed representations and structures that linearly propagate
information across time) are profoundly different from those
postulated in linguistics (such as categorical labels and tree
structures). Still, as we will see below, for all the enthusiasm for
LODNA within NLP, this line of work is hardly having any impact on the
current debate in theoretical linguistics.

In this chapter, after introducing, as an example of LODNA, the by-now
``classic'' domain of long-distance agreement probing (Section
\ref{sec:lda}), I will present evidence for the claim that this sort
of research, despite the intriguing patterns it uncovers, is hardly
affecting contemporary linguistics (Section \ref{sec:thegap}). I will
then argue that this gap stems from lack of clarity about its theoretical
significance (Section \ref{sec:theoretical}). In particular, I will
show that modern deep networks cannot be treated as blank slates meant
to falsify innateness claims. They should rather be seen as
algorithmic linguistic theories making predictions about utterance
acceptability. I will however outline several issues that are currently
standing in the way of taking deep nets seriously as linguistic
theories. I will conclude in Section \ref{sec:conclusion} by briefly
discussing why taking this stance might be beneficial to computational
and theoretical linguistics, and by sketching two possible ways to
pursue LODNA-based linguistic theorizing.

\section{Linguistic-oriented analysis of deep nets: The case of long-distance agreement}
\label{sec:lda}

Linguists identify sensitivity to syntactic structure that is not
directly observable in the signal as one of the core properties of
human grammatical competence \citep{Everaert:etal:2015}. A
paradigmatic test for structure sensitivity comes from agreement
phenomena. For example, subject-verb number agreement in an English
clause depends on the c-command relation between the subject noun
phrase and the corresponding verb, and it is not affected by nouns
intervening between the NP head and the verb:

\ex. [The \textbf{kid} [near the \emph{toys} in the \emph{boxes}]] \textbf{is} tired.\label{ex:nda}

In \ref{ex:nda}, an example of \emph{long-distance agreement}, the fact that two plural nouns (\emph{toys} and
\emph{boxes}) directly precede the main verb \textit{is} does not
affect its number, as the only noun that entertains the right relation
with the verb is \emph{kid}. As Everaert and colleagues'
(\citeyear{Everaert:etal:2015}) motto goes, it's all about
\emph{structures}, not strings!

Current deep network architectures, such as
long-short-term memory networks \citep[LSTMs,][]{Hochreiter:Schmidhuber:1997}, convolutional
networks \citep[CNNs,][]{Kalchbrenner:etal:2014} or Transformers \citep{Vaswani:etal:2017}, do not encode (at least, by conscious designer decision) any
prior favoring a structural analysis of their input over a sequential one. It is natural then to ask whether they
are able to correctly handle structure-dependent
phenomena, such as long-distance agreement. Consequently, starting
with the influential work of \citet{Linzen:etal:2016},
long-distance agreement has become a standard test to probe their linguistic abilities.

Probably the most thorough analysis of long-distance number agreement
in deep networks was the one we carried out in
\citet{Gulordava:etal:2018}. We focused on LSTMs trained as
\emph{language models}. That is, the networks were trained by exposing
them to large amounts of textual data (samples from the Wikipedias of
the relevant languages), with the task of predicting the next word
given the context. No special tuning for the long-distance agreement
challenge was applied. After this generic training, the networks were
presented with sentence prefixes up to and excluding the second item
in an agreement relation (e.g., \emph{The kid near the toys in the
  boxes\ldots}), and the probability they assigned to continuations
with the right or wrong agreement (\emph{is/are}) was measured. The
experiment was conduced with a test set of genuine corpus examples in
4 languages (English, Hebrew, Italian and Russian), and considering
various agreement relations (not only noun-verb but, also, for
example, verb-verb and adjective-noun). The networks got the right
agreement with high accuracy in all languages and for all
constructions.

Even more impressively, the networks were also able to get
 agreement right when tested with nonsense sentences such as the one in
\ref{ex:nonsense}, showing that they must extract
syntactic rules at a rather abstract level.

\ex. The \textbf{carrot} around the \emph{lions} for the \emph{disasters}\ldots \textbf{sings/*sing}.\label{ex:nonsense}

Finally, we compared the agreement accuracy of
the Italian network with that of native speakers (both on
corpus-extracted and nonsense sentences), finding that the network is
only marginally lagging behind human performance.

Other studies tested different deep architectures, such as CNNs
\citep{Bernardy:Lappin:2017} and Transformers \citep{Goldberg:2019b},
confirming that they also largely succeed at long-distance agreement.

Deep nets have been tested for a number of other linguistic
generalizations, such as those pertaining to filler-gap constructions,
auxiliary fronting and case assignment. See \citet{Linzen:Baroni:2020}
for a recent survey of LODNA specifically aimed at
linguists.\footnote{\citet{Lappin:2021} also provides a review of some
  of the relevant work, as part of a book-length treatment of the
  linguistic and cognitive implications of deep learning models.} In pretty much all cases, while they departed here and
there from human intuition, deep nets captured at
least the general gist of the phenomena being investigated.



\section{The gap}
\label{sec:thegap}

Results such as the ones on long-distance agreement I briefly reviewed
should provide food for thought to theoretical linguists, since, as already
mentioned, deep nets ostensibly possess very different priors from those
postulated by linguists as part of the universal language faculty,
such as a predisposition for hierarchical structures
\citep{Hauser:etal:2002,Berwick:Chomsky:2016,Adger:2019}. In reality,
however, the growing body of work on LODNA is almost completely
ignored in the current theoretical linguistics debate.

To sustain this claim with quantitative evidence, I looked at the
impact of Tal Linzen's original paper on long-distance agreement in
deep nets \citep{Linzen:etal:2016}. This is a highly-cited paper,
having amassed 514 Google Scholar citations in less than 5
years.\footnote{Google Scholar queried on May 27th, 2021.} I sifted
through these citations, keeping track of how many came from
theoretical linguistics (under a very broad notion of what counts as
theoretical linguistics). I found that only 6 citations qualified. Of
these, 3 were opinion pieces, one of them written by Linzen himself. Note that the
article does not lack general interdisciplinary appeal, as shown by many
citations from psycho- and neuro-linguistics, and even 4 citations
from the field of computational agricultural studies!

Perhaps Google Scholar does a poor job at tracking theoretical
linguistic work. Indeed, David Adger's recent \emph{Language
  Unlimited} volume \citep{Adger:2019} does extensively discuss
Linzen's article, but I did not find it among the studies citing it
according to Scholar. Thus, as a supplementary source of evidence, I
also downloaded all papers from the front page of LingBuzz, a popular
linguistics preprint archive.\footnote{Papers downloaded from
  \url{https://ling.auf.net/lingbuzz} on May 27th, 2021. I downloaded
  all \emph{freshly changed} and \emph{new} papers, as well as all the
  papers in the \emph{Top Recent Downloads} and \emph{Last 6 months}
  sections of the front page.} I filtered out papers that do not
qualify as theoretical linguistics. Again, I tried to be inclusive:
I excluded, for example, one paper about the aftermath of the ``Pinker
LSA letter'' controversy \citep{Kastner:etal:2021}, but I did include
one about phonosymbolysm in Pok\'{e}mon character names
\citep{Kawahara:etal:2021}. This left me with a corpus of 37 papers. I
then went through their bibliographies, looking for references to
deep learning work, and finding\ldots none!\footnote{I had performed a
  similar experiment in March 2021, by collecting papers from the
  latest issues of \emph{Linguistic Inquiry}, \emph{Natural Language
    and Linguistic Theory} and \emph{Syntax}, with the very same
  outcome (no reference whatsoever to deep learning work).}

It is not fair to impute this lack of references to a putative endogamous bent of theoretical linguistics. To the contrary, the papers in my mini-corpus reveal considerable interdisciplinary breadth, with frequent references
to neuroscience, ethology, psycholinguistics and sociolinguistics;
they include statistical treatments of experimental and corpus data;
and they use sophisticated computational tools, such as
graph-theoretical methods. It is really NLP, and in particular
deep-learning-based NLP, that is missing from the party.

To understand this gap, we need to ask: why should linguists care
about the grammatical analysis of deep networks? What is it supposed
to tell us about human linguistic competence? In other words, what is
the theoretical significance of LODNA?

\section{The theoretical significance of linguistically-oriented deep net analysis}
\label{sec:theoretical}

When LODNA researchers situate their work within a broader theoretical
context, it is invariably in terms of nature-or-nurture arguments
resting on a view of deep nets as blank slates. For example, when
asked about the significance of his work for theoretical linguistics,
Tal Linzen told me that deep-net simulations ``can help linguists
focus on the aspects [\ldots] that truly require explanation in terms
of innate constraints. If the simulation shows that there is plenty of
data for the learner to acquire a particular phenomenon, maybe there's
nothing to explain!''  (Tal Linzen, p.c.).

Similar claims are sprinkled throughout LODNA papers. Here are just a
few examples (from otherwise excellent papers): ``Our results also
contribute to the long-running nature-nurture debate in language
acquisition: whether the success of neural models implies that
unbiased learners can learn natural languages with enough data, or
whether human abilities to acquire language given sparse stimulus
implies a strong innate human learning bias''
\citep{Papadimitriou:Jurafsky:2020}.  ``The APS [(argument from the
poverty of the stimulus)] predicts that any artificial leaner trained
with no prior knowledge of the principles of syntax [\ldots] must fail
to make acceptability judgments with human-level accuracy. [\ldots] If
linguistically uninformed neural network models achieve human-level
performance on specific phenomena [\ldots], this would be clear
evidence limiting the scope of phenomena for which the APS can hold''
\citep{Warstadt:etal:2019}. ``[I]f such a device [(a neural network)]
could manage to replicate fine-grained human intuitions inducing them
from the raw training input this would be evidence that exposure to
language structures [\ldots] should in principle be sufficient to
derive a syntactic competence, against the innatist hypothesis''
\citep{Chowdhury:Zamparelli:2018}.

\subsubsection*{Deep nets are linguistic theories, not blank slates}

If blank slate arguments were (perhaps) valid when looking at the
simple connectionist models of the eighties
\citep{Rumelhart:etal:1986,Churchland:1989,Clark:1989}, all modern
deep networks possess highly-structured innate architectures that
 considerably weaken them.  Consider, for example, the Transformer
\citep{Vaswani:etal:2017}, the current darling of NLP. A Transformer network is structured into a number of
layered modules, each involving a complex bank of linear and
non-linear transformations. These, in turn, differ in profound ways
from the innate structure of a LSTM
\citep{Hochreiter:Schmidhuber:1997}. For example, a LSTM will read a
sentence one token at a time, and will use a recurrent function to
preserve information across time, whereas the Transformer will read
in the whole sentence at once, and use an extended backward and
forward attention system to incorporate contextual information.

Even more importantly, as demonstrated by the widespread interest of
NLP and machine learning researchers in proposing new architectures,
differences in the supposedly ``weak'' and ``general'' biases of
different deep nets lead them to behave very differently, given the
same input data.

A striking illustration of this was recently provided by
\citet{Kharitonov:Chaabouni:2021} in a study of so-called \emph{seq2seq} deep nets, that is, networks trained to associate input and output sequences (as
in, e.g., a translation task).

Kharitonov and Chaabouni trained such
networks on really tiny corpora that severely underspecify the
input-output relation. The test-time behavior of the network in cases
where different generalizations lead to different outputs was then
inspected, to reveal which innate preferences the networks brought to
the task.

In one of their experiments, the whole training corpus
consists of the following \emph{input} $\rightarrow$ \emph{output}
examples.

\ex. aabaa $\rightarrow$ b\\
     bbabb $\rightarrow$ a\\
     aaaaa $\rightarrow$ a\\
     bbbbb $\rightarrow$ b\label{ex:mini-corpus}

The mini-corpus in \ref{ex:mini-corpus} is compatible with (at
least) two rules: a ``hierarchical'' one, stating that the output
is generated by taking the character in the middle of the input;
and a ``linear'' generalization, stating that the output is the
third character in the input sequence.\footnote{This can be seen
  as a schematic reproduction of classic
  poverty-of-the-stimulus thought experiments, such as the one built around
  English auxiliary fronting by \citet{Chomsky:1968}.}

After training it with just the examples in \ref{ex:mini-corpus}, a
network is exposed to a new input where the two rules lead to
different predictions, e.g., \emph{aaabaaa}, where the hierarchical
generalization would pick \emph{b} and the linear one \emph{a}.

Of four widely-used seq2seq models tested by Kharitonov and Chaabouni,
two (LSTMs with attention and Transformers) show a strong preference
for the hierarchical generalization, and two (LSTMs without attention
and CNNs) show a strong preference for the linear
generalization.

Studies such as this invalidate any blank-slate claim about deep
nets. 
It is more appropriate, instead, to look at deep nets as
\emph{linguistic theories}, encoding non-trivial structural priors
facilitating language acquisition and processing. More precisely, we
can think of a deep net architecture, before any language-specific
training, as a general theory defining a space of possible grammars,
and of the same network trained on data from a specific language as a
\emph{grammar}, that is, a computational system that, given an input
utterance in a language, can predict whether the sequence is
acceptable to an idealized speaker of the language
\citep[e.g.,][]{Chomsky:1986,Sag:etal:2003,Mueller:2020}.\footnote{Just like in linguistics
  \citep[e.g.,][]{Murphy:2007,Lau:etal:2017,Spouse:Schuetze:2019},
  there is considerable debate on the best way to elicit acceptability
  judgments from trained deep net models in order to compare them to human data, and on
  whether such judgments should be probabilistic or categorical
  \citep[e.g.,][]{Linzen:etal:2016,Chowdhury:Zamparelli:2018,Warstadt:etal:2019,Niu:Penn:2020}.} %

It is undoubtedly easier to inspect the inner workings of a symbolic
linguistic theory than those of a trained deep net, and indeed a
classic objection against artificial neural networks as cognitive
theories is that they are unopenable black boxes
\citep[e.g.,][]{McCloskey:1991}. However, going hand in hand with the development of more complex
models, the field has also made extensive progress in the development
of methods to analyze their states and behaviors
\citep{Belinkov:Glass:2019}, providing strong methodological support
for a systematic analysis of deep nets.

Why don't we see, then, many articles positioning deep nets as alternative
or complementary theories to traditional grammatical formalism? I
believe that two crucial ingredients are still missing, before deep
nets can seriously contribute to contemporary linguistic theorizing.


\subsubsection*{The problem of low commitment to models}

Differences between deep nets, as we have discussed above, are
huge. \emph{Mutatis mutandis}, the difference between an LSTM, reading
an input token at a time and building a joint representation through its
recurrent state, and a Transformer, processing all input tokens in parallel
to create multiple context-weighted representations, might be as large
as that between a derivational and a constraint-based theory in formal
linguistics.\footnote{At a deep mathematical level, recurrent networks (such as LSTMs) and Transformers might be more related than what a superficial comparison might suggest \citep[see, e.g.,][]{Katharopoulos:etal:2020}, just like some differences between derivational and constraint-based grammars might be more apparent than substantive \citep{Hunter:2019}.} 

And, yet, researchers investigating the linguistic behavior of these
architectures almost never provide a theoretically grounded motivation
for why they focused on one architecture or the other. Interest tends
to shift with the state of the art in applied tasks such as machine
translation or natural language inference. So, if nearly all early
LODNA papers focused on recurrent LSTM networks, nowadays the field has nearly
entirely shifted to analyzing Transformer networks, not because the
latter were found to be more plausible models of human language
processing (if anything, their ability to read and process massive
windows of text in parallel makes them \emph{less} plausible models than
recurrent networks), but because they became the mainstream approach in
applied NLP, thanks to their astounding performance in applied tasks.

As a concrete illustration of this phenomenon, we can compare the LODNA
papers from the first
(2018) and third (2020) editions of the BlackBox NLP workshop (one of
the core events in the
area).\footnote{\url{https://www.aclweb.org/anthology/volumes/W18-54/};\\
  \url{https://www.aclweb.org/anthology/volumes/2020.blackboxnlp-1/}}
In the 2018 edition, I found 13 full papers that broadly qualify as LODNA. Of them, 12 focus on LSTM analysis, with the
remaining one already looking at the Transformer. In two years, the
balance has completely shifted. All 9 relevant papers in the 2020
editions analyze some variant of the Transformer, with two also
including LSTM variants among the comparison models. Importantly, in
none of these papers there is a linguistically-oriented (or even
engineering-oriented) discussion of \emph{why} the Transformer was
picked over the LSTM or other architectures. Indeed, in a few cases,
earlier work that was based on LSTMs is cited as corroborating
evidence, only mentioning in passing that it was based
on a (profoundly different) architecture.

The problem is mostly sociological: NLP puts a strong (and
reasonable) emphasis on whichever models work best in applications, and
consequently analytical work will also tend to concentrate on such
models. However, if radical changes in the underlying architecture are
not motivated by linguistic considerations, and indeed they tend to be
completely glossed over, it is hard to take this work seriously
from the perspective of linguistic theorizing.\footnote{There are
  important exceptions. Work that does put an emphasis on the
  linguistic motivation of architectural choices includes that of
  Chris Dyer and colleagues on recurrent neural network grammars
  \citep[e.g.,][]{Kuncoro:etal:2018a}, and that of Paul Smolensky and
  colleagues on tensor product decomposition networks
  \citep[e.g.,][]{McCoy:etal:2019b}.}

\subsubsection*{Lack of mechanistic understanding}

A good linguistic theory should not only fit what is already known about a
language, but also make predictions about previously unexplored
patterns. This is the typical \emph{modus operandi} in
formal syntax, where, for example, hypotheses about possible syntactic
configurations lead to strong typological predictions about
acceptable adverb and adjective orders
\citep[e.g.,][]{Cinque:1999,Cinque:2010}.

The standard approach in LODNA, instead, is to check whether pre-trained models
capture well-known patterns, such as vanilla English
subject-verb number agreement. The occasional focus on cases outside the
standard paradigm is typically meant to highlight obviously \emph{wrong}
predictions made by the model \citep[e.g.,][show that, in some syntactic configurations, LSTMs let the verb agree with the first noun
in a sentence even if it is not its
subject]{Kuncoro:etal:2018b}.

What we are doing, then, is an extensive (and important!)
sanity check of our systems, rather than using them to widen the
coverage of linguistic phenomena we are able to explain through
computational modeling.

In order to move from sanity checks to prediction generation, we need
however to have a good-enough understanding of how the underlying
mechanisms implemented in a network cause their linguistic
behavior. The large majority of LODNA studies focuses on the
behavioral level. We need to shift the focus to a mechanistic
\emph{neural} level, so to speak.

As an example of the kind of study combining a granular understanding
of a model's inner workings with a non-trivial prediction tested in
humans, I will briefly summarize the detailed analysis of deep net long-distance number agreement that we reported
in \citet{Lakretz:etal:2019,Lakretz:etal:2021}.

In the first of these studies, a cell-by-cell analysis of pre-trained LSTMs performing the
subject-verb agreement task revealed that they develop a
sparse mechanism to store and propagate a single number feature between
subject and verb. This sparse grammar-aware circuit is
complemented by a distributed system that can fill in the number
feature based on purely sequential heuristics.

This leads to an interesting prediction for sentences with two embedded long-distance dependencies, such as:

\ex. \label{ex:multiple-embeddings}
 The \textbf{kid}$_1$ that the \textbf{dogs}$_2$ near the \emph{toy} \textbf{like}$_2$\textbf{/*likes}$_2$ \textbf{is}$_1$\textbf{/*are}$_1$ tired.

Here, the sparse grammar-aware mechanism will be activated when
\emph{kid} is encountered, and, due to its sparsity, it will not be able to also record the
number of \emph{dogs}. Consequently, once \emph{like(s)} is
encountered, the heuristic distributed system will take over, and
it will wrongly predict the singular form, since the sequentially closer
noun is \emph{toy}. On the other hand, once \emph{is/are} is
reached, the feature stored in the sparse long-distance circuit can
be released, correctly predicting a preference for singular
\emph{is}. This is an interesting prediction because,
intuitively but contrary to it, the longer distance \emph{kid-is} relation
should be harder to track than the shorter-distance one 
connecting \emph{dogs} and \emph{like}.

In \citet{Lakretz:etal:2021}, we proceeded to test the prediction both in
LSTMs and with human subjects. We did indeed find the predicted
inner/outer agreement asymmetry both in machines and (more weakly) in
humans. This suggests that agreement might be
implemented by means of sparse feature-carrying mechanisms in humans
as well.


Lakretz' study took about 4 years to run. By the time it was
completed, it presented a detailed analysis of a model, the LSTM, that
many in NLP would find obsolete. Its focus on a single
grammatical construction might look quaint, now that the field has
moved towards large-scale evaluation suites probing models on a
variety of phenomena and tasks
\citep[e.g.,][]{Conneau:Kiela:2018,Conneau:etal:2018,Marvin:Linzen:2018,Wang:etal:2019,Warstadt:etal:2019}. Yet,
if we want to reach the sort of understanding of a deep model's
inner working that can be useful to gain new insights on human linguistic
competence and behavior, I argue that we should have more studies
running at the same slow, thorough, narrow-focused pace of this
project.\footnote{Independent progress in causal intervention methods applied to modern language models \citep[e.g.,][]{Meng:etal:2022} will hopefully speed up and generalize the process of understanding the mechanics of these deep nets.}

\section{Conclusion}
\label{sec:conclusion}

Language models based on deep network architectures such as the LSTM
and the Transformer are computational devices that, by being exposed
to large amounts of natural text, learn to assign probabilities to
arbitrary word sequences. In the last five years or so, a rich
tradition of studies has emerged that analyzes such models in order to
understand what kind of ``grammatical competence'' they possess.

The results of these studies are often intriguing,
revealing the sophisticated linguistic skills of deep nets, as well as
interesting error patterns. However, such studies have had very little
impact on theoretical linguistics.

I attributed this gap to the fact that these studies tend to lack
a clear theoretical standing and, when they do, it is one based on the
wrong idea that we should treat modern deep nets as \emph{tabulae
  rasae} without strong innate priors. Deep nets do possess such
innate priors, as shown by the fact that different models trained on
the same data can extract dramatically different
generalizations. %
I proposed that a more solid theoretical standing for
 the linguistic analysis of deep nets can be achieved by treating them as
\emph{algorithmic linguistic theories}.

I discussed above some concrete roadblocks we must overcome if we want
to seriously adopt this stance. I will conclude by
briefly explaining why I think that such a stance is beneficial for both
computational and theoretical linguists, and by providing quick
sketches of what deep-net-based linguistic theorizing could
look like.

\subsubsection*{Why should computational linguists care?}

The incredible progress in deep learning for NLP we've
seen in the last few years must be entirely credited to NLP and machine-learning
practitioners interested in solving concrete challenges such as
machine translation. Ideas from theoretical linguistics have played no
role in the area \citep{Lappin:2021}, and there is no clear reason, in
turn, why computational linguists interested in practical NLP
technologies should care about the implications of their work for
linguistics.

However, the success of events such as the already mentioned
Society for Computation in Linguistics conference and BlackBox NLP workshop,
as well as the fact that all major NLP
conferences now feature special tracks on linguistic analysis of
computational models, suggest that there is a significant
sub-community of computational linguists who \emph{are} interested in
the linguistic implications of deep learning models. 

These researchers should be bothered by the fact that their work is
not having an impact on mainstream theoretical linguistics. Clarifying
the theoretical status of deep net simulations, and in particular boldly
presenting them as alternative linguistic theories,
might finally attract due attention from the linguistics
community.

\subsubsection*{Why should theoretical linguists care?}

Deep nets attained incredible empirical results in tasks that heavily
depend on linguistic knowledge, such as machine translation
\citep{Edunov:etal:2018}, well beyond what was ever achieved by
symbolic or hybrid systems. While it is possible that deep nets are
relying on a completely different approach to language processing than
the one encoded in human linguistic competence, theoretical linguists
should investigate what are the building blocks making these systems
so effective: if not for other reasons, at least in order to explain
why a model that is supposedly encoding completely different priors
than those programmed into the human brain should be so good at
handling tasks, such as translating from a language into another, that
should presuppose sophisticated linguistic knowledge.

I conjecture however that deep nets and traditional symbolic theories
are both valid algorithmic approaches to modeling human linguistic competence,
and that they are complementary in the aspects they best explain. The
more algebraic features of language, such as recursive structures, are
elegantly handled by traditional linguistic formalisms such as
generative syntax \citep{Mueller:2020} and formal semantics
\citep{Heim:Kratzer:1998}. However, language has other facets, in
particular those where the fuzzy, large-scale knowledge that
characterizes the lexicon is involved, where such theories
struggle. Neural language models, by inducing a large set of
context-dependent and fuzzy patterns from natural input, and by being
inherently able to probabilistically generate and process text, should
be better equipped to handle phenomena such as polysemy, the partial
productivity of morphological derivation, non-fully-compositional
phrase formation and diachronic shift
\citep[e.g.,][]{Marelli:Baroni:2015,Vecchi:etal:2017,Lenci:2018,Boleda:2020}.\footnote{These
  references mostly discuss a precursor of neural language models
  known as \emph{distributional semantics}, but the same accounts
  could be replicated and extended using latest-generation neural
  language models.}

From this angle, the current emphasis of LODNA on exactly those
phenomena (such as long-distance agreement) that are already
satisfactorily captured by traditional algebraic models might be
misguided. Curiously, even staying within the domain of syntax,
there is no work I am aware of focusing instead on those patterns,
such as partially lexicalized constructions
\citep[e.g.,][]{Goldberg:2005,Goldberg:2019}, where the fuzzier rules
typically learned by neural networks might give us novel insights into
human generalization.

\subsubsection*{Do neural network theories require a switch from algebraic to distributed models of linguistic competence?}

The main topic of this volume is the role of algebraic systems in the
representation of linguistic knowledge. By proposing a trained
Transformer, with its billions of weights and its continuous
activation vectors, as a ``linguistic grammar'', I am \emph{de facto}
implying that the appropriate level to represent linguistic knowledge
is not algebraic, but massively distributed. This requires a radical
methodological shift in the way linguistic models are
studied. Standard rule- or constraint-based systems can easily be
probed by direct inspection. With deep networks, model probing
requires sophisticated experiments, of the kind that the LODNA
literature has only partially started designing, especially in terms
of understanding the causal mechanisms underlying a model's linguistic
behavior.

However, I would like to leave the issue of the right level for
deep-net-based linguistic theorizing open. Optimality Theory
\citep{Prince:Smolensky:2004} was the most fruitful outcome of early
attempts to bring together linguistics and connectionism. Optimality
Theory is an algebraic approach whose principles are inspired by how
linguistic constraints might be implemented by a traditional neural
network. Could the way in which LSTMs or Transformers process
linguistic information similarly inspire a symbolic theory of
language? Perhaps, one that is not based on tree structures but on
storage and retrieval mechanisms akin to gating and attention?
\\
\\
To conclude, despite the criticism I vented to some aspects of the
field, I think that LODNA is one of the most exciting things that has
happened to cognitive science in the last six years. I hope that,
once we clarify its theoretical standing, and as we deepen our
understanding of how deep networks accomplish linguistic tasks, the
body of evidence assembled in this area will finally have the impact
it deserves on linguistics at large.

\section*{Acknowledgments}

I would like to thank the anonymous reviewer, Jelke Bloem, Grzegorz
Chrupała, Ido Dagan, Roberto Dess\`{i}, Emmanuel Dupoux, Dieuwke
Hupkes, Shalom Lappin, Yair Lakretz, Paola Merlo, the members of the
UPF Computational Linguistics and Linguistic Theory group, the
participants in the EACL 2021 Birds-of-a-Feather Meetup on Linguistic
Theories, the audience at EACL 2021 and, especially, David Adger,
Gemma Boleda, Roberta D'Alessandro, Chris Dyer, Tal Linzen, Louise
McNally, Tom McCoy, Paul Smolensky and Adina Williams for a mixture of
advice, stimulating discussion and constructive feedback.

\end{document}